\documentclass[twocolumns]{IEEEtran}
\usepackage{balance}
\usepackage{cite}
\usepackage{url}
\usepackage[utf8]{inputenc}
\usepackage{booktabs}
\usepackage{graphicx}
\usepackage{tabularx}
\usepackage{multirow}
\usepackage[font=small]{caption}
\usepackage[hyphenbreaks]{breakurl}
\usepackage{textcomp}
\usepackage{xcolor}
\usepackage{diagbox}
\usepackage{subfigure}
\usepackage{float}
\usepackage{amsmath,amssymb,amsfonts}
\usepackage[hidelinks]{hyperref}
\usepackage[hyphenbreaks]{breakurl}
\usepackage{url}

\makeatletter

\newcommand{\Rmnum}[1]{\expandafter\@slowromancap\romannumeral #1@}
\makeatother

\usepackage[textsize=tiny, textwidth=1.5cm]{todonotes}
\usepackage{xcolor, soul}
\sethlcolor{yellow}

\usepackage{listings}
\AtBeginDocument{%
  \providecommand\BibTeX{{%
    \normalfont B\kern-0.5em{\scshape i\kern-0.25em b}\kern-0.8em\TeX}}}

\begin{document}

\title{Generative Anomaly Detection for Time Series Datasets}

\author{
    \IEEEauthorblockN{Zhuangwei Kang\IEEEauthorrefmark{1}, Ayan Mukhopadhyay\IEEEauthorrefmark{1}, Aniruddha Gokhale\IEEEauthorrefmark{1}, Shijie Wen\IEEEauthorrefmark{2}, Abhishek Dubey\IEEEauthorrefmark{1}} \\
    \IEEEauthorblockA{\IEEEauthorrefmark{1} Institute for Software Integrated Systems, 
Vanderbilt University}
    \IEEEauthorblockA{\IEEEauthorrefmark{2} Cisco Systems, Inc.}
}
\date{June 2022}

\maketitle



\textcolor{red}{A version of this paper was presented at the ITSC 2022}

\section{Introduction}
In many real-world monitoring scenarios, such as stock returns, road traffic conditions, data center KPIs, and personal health metrics, collected data appears in the form of multivariate time series (MTS). Monitoring is critical in practice. For instance, Transportation Management Centers are critical for managing the surface road network; delays accrued during the monitoring phase delay response and resolution~\cite{yasas2021}. Frequently, secondary crashes and long-clearance times lead to additional congestion on critical arterial road segments. To improve the real-time monitoring of extensive road networks, transportation agencies are increasing the available sensing modalities, often in smart corridors. However, this drastic increase in the number of sensors raises an essential question from an operational perspective---how can transportation agencies monitor thousands of sensors in (near) real-time to detect incidents of interest? Our conversations with local transportation agencies revealed that this monitoring is largely performed manually, an infeasible strategy in the long run. One approach to enable transportation agencies to utilize an extensive array of sensors is to detect potentially anomalous patterns in real-time using the data generated by the sensors; then, human experts (or potentially decision-theoretic approaches~\cite{pettet2022designing}) can narrow their focus on the anomalies and take necessary operational actions. Similarly, in large data centers, server machine KPIs, such as CPU, memory, TCP, UDP metrics, are periodically collected as multi-variate time series by some profilers. These KPIs are critical for estimating machine health status and making timely responses. 

\textbf{Challenges} Anomaly detection is a rich field in data mining and has been explored widely in the domain of transportation networks. While detection has been traditionally done for various traffic condition variables with techniques such as CUSUM~\cite{wilbur2019decentralized}, K-nearest Neighbors~\cite{harrou2020traffic}, Isolation Forests~\cite{mercader2020automatic}, and forecasting models (e.g., ARIMA~\cite{moayedi2008arima}), deep neural networks (DNN) have gradually become the state-of-the-art due to the remarkable capability of modeling high-dimensional MTS data. However, despite the universal approximation power of DNN on learning unknown data distributions, performing anomaly detection on MTS is still challenging. For example, many DNN-based approaches either rely on an uncontaminated training dataset to learn the normal traffic patterns (semi-supervised) or reframe the detection task as a classification task using a fully-labeled traffic mobility dataset (supervised).

Such approaches, however, are not practical in general, and moreover, such data often do not account for a large number of incidents such as phantom traffic jams, slowdowns, and weather hazards. Further, even if labels are available, such classifiers and approaches identify point anomalies, where the observation is clearly far away from what is expected globally. In addition, traditional anomaly detection techniques focus on maximizing the accuracy of detection. However, in the specific use case of transportation centers, the goal of such a detector is to ensure that the search space for monitoring is shrunk for domain experts. As a result, in practice, the detector must demonstrate high recall with relatively low precision, i.e., false negatives are more costly than false positives. Finally, an additional challenge is proactive model improvement; agencies must ensure that the learned model used to detect anomalies is improved proactively to detect potentially unseen anomalies. 

\textbf{Contributions} This work systematically addresses these challenges by developing a multi-variate anomaly detection framework based on conditional normalizing flow, a probabilistic generative model that can tractably perform density estimation and sampling in extremely high dimensional spaces~\cite{rezende2015variational}. Through this approach, we can model the multimodal distributions of time series data. In particular, we summarize the contributions of this paper as follows:
\begin{enumerate}
    \item We propose a principled MTS anomaly detection and diagnosis model for traffic data that comprises an LSTM-Encoder-Decoder (LSTM-EncDec) model and a Normalizing Flow architecture~\cite{rezende2015variational}, specifically a RealNVP~\cite{dinh2016density} flow. The former makes sequence-to-sequence forecasting with a sliding-window scheme to extract internal spatial-temporal information from  ground-truth data. The flow model is used to model complex data distribution in the high-dimensional transit data. Specifically, it performs conditional density estimation using the outputs of the forecasting model. 
    \item To ensure tractability of our approach, we divide the road network of a city into clusters, and perform anomaly detection at the granularity of clusters.
    \item Then, we use a simpler density estimator based on a kernel density function to identify anomalies at the granularity of features(e.g., road segments in transportation systems or KPIs for server machines).
    \item We compare our approach with existing state-of-the-art baselines using traffic data collected from the City of Nashville, Tennessee as well as an open dataset on server machines. Experimental results show that our approach has superior performance and sensitivity on anomaly detection in traffic networks.
\end{enumerate}
\section{Background}
\label{sec:background}
In this section, we provide some preliminary knowledge about general normalizing flows and RealNVP.

\subsection{Normalizing Flow}
Normalizing flows define a series of bijective transformations that can transform the probability density $p_{X}(x)$ of a random variable $X \in \mathbb{R}^D$ to a well-known base distribution $p_{Z}(z)$ defined by a random variable $Z \in \mathbb{R}^D$~\cite{kobyzev2020normalizing}. The random variable $Z$ is chosen such that it has an explicit probability density function. The problem of training the normalizing flow is to learn an invertible transformation, $f$ such that $z = f(x)$ and $x  = f^{-1}(z)$. The transformation is a sequence of bijective functions composed together, i.e., $f=(f_1.f_2\cdots$).
Once learned, the forward mapping, $X \rightarrow Z$ can be used for density estimation and the inverse mapping $Z \rightarrow X$ can be used for sampling (synthetic data generation). This mapping presents a key advantage that enables exact density estimation without loss of dimensional information, making it suitable for anomaly detection. 
In particular, the marginal likelihood $p_{X}(x)$ can be expressed as:
\begin{equation}\label{eqn:change_of_variable}
\begin{aligned}
    p_{X}(x) &= p_{Z}(f(x)) \left| \det\left(\frac{\partial f(x)}{\partial x}\right)\right|
\end{aligned}
\end{equation}
where $p_{Z}(f(x))$ is density of $x$ under the base distribution $p_{Z}$ and $\det\left(\frac{\partial f(x)}{\partial x}\right)$ is the determinant of the Jacobian of $f$. The main challenges of modeling arbitrary distributions using normalizing flow lie in designing the compositional and invertible transformation $f$. Further, the choice of the architectures are restricted by the need for the efficient computation of the determinant of the Jacobian matrix.

\subsection{RealNVP}\label{subsec:realnvp}
One of the recent innovations in normalizing flow is the use of the real-valued non-volume preserving transformations~\cite{dinh2016density} as the function $f$. Effectively, RealNVP is a set of affine coupling layers, one of the possible bijective transformations that can be used to design the composition $f$. To explain this further, consider the example of a single layer transformation (several such layers are composed in practice) $Y$ that maps $X$ to $Z$. RealNVP transformation $Y$ partitions $X$ into two disjoint groups, where the first $d$ dimensions remain unchanged while the latter part, i.e., from the $d+1$-th to the $D$-th dimension, undergoes an affine transformation. Formally, 
\begin{align}\label{eqn:transformation}
\begin{split}
    y^{1:d} &= x^{1:d} \\
    y^{d+1:D} &= x^{d+1:D} \odot \exp(s_{net}(x^{1:d})) + t_{net}(x^{1:d})
\end{split}
\end{align}
$s_{net}$ and $t_{net}$ indicate a ``scale'' and a ``translation'' function respectively and $\odot$ stands for element-wise product. The representation power of RealNVP depends on $s_{net}$ and $t_{net}$, which can be any arbitrarily complex function (often a neural network architecture). Note that because the first $d$ dimensions remain unchanged during transformation, to make the flow model capture the full picture of input space, RealNVP swaps active and inactive dimensions in an alternating manner. A convenient way to realize this is to multiply the $D$ dimensional inputs and outputs with a binary mask vector. 

RealNVP guarantees its computation of Jacobian function is efficient because the Jacobian is a block-triangular matrix, where elements on the diagonal are an identity matrix and a diagonal matrix whose diagonal elements correspond to the vector $\exp\left(s_{net}(x^{1:d})\right)$. Therefore, the determinant of Jacobian, which simplifies to $\exp(\sum_j(s_{net}(x^{1:d})_j))$ can be efficiently computed. If the flow is implemented using $K$ such layers, which is required to ensure better learning, the  probability density of a given sample $x$ can be calculated as follows: 
\small
\begin{align}\label{eqn:nvp_density_estimation}
\begin{split}
    log(p_{X}(x)) = \log(p_{Z}(z))
    +\sum_{k=1}^K \log\left( \left| \exp(\sum_j(s_{net}^k(y_{k-1}^{1:d}))_j) \right| \right)
\end{split}
\end{align}
\normalsize
where the first term denotes the likelihood of $z$ (transformed from $x$) on the base distribution, and the second term represents the accumulated changes while transforming $x$ to $z$. Thus, the training objective of RealNVP is to find the right set of hyperparameters of the $s_{net}$ and $t_{net}$ that maximize the overall likelihood of the observed data, which can be denoted as 
\begin{align}
  \theta^{*} = \underset{
    \theta \in \Theta}{\arg\max} \frac{1}{|\mathcal{D}|} \sum_{x\in \mathcal{D}} \log p_X(x; \theta)
\end{align}
, where $\mathcal{D}$ is the observed data and $\theta$ denotes parameters of $s_{net}$ and $t_{net}$ functions.

\section{Methodology}
\label{sec:method}
The problem statement of MTS anomaly detection and the function and principle of each component in our framework (as shown in figure~\ref{fig:model}) are concentrated and discussed in this section.

\begin{table}[]
\centering
\setlength\tabcolsep{3pt}
\begin{tabularx}{\columnwidth}{l|p{6.8cm}}
\hline
Symbol            & Description                                                                   \\ \hline
$\mathcal{D}$     & the traffic congestion dataset                                                \\ \hline
$X$               & set of congestion rate time series                                            \\ \hline
$T$               & total length of the traffic dataset                                           \\ \hline  
$i, t$            & a specific roadway segment and time step                                     \\ \hline
$\lambda^{t}$     & a time feature vector for time t                                              \\ \hline
$S_i$             & the \textit{i}th road segment                                                 \\ \hline
$x_{i}$           & univariate time series, congestion rate of the \textit{i}th road segment      \\ \hline
$x^t$             & a vector of congestion values at time t                                       \\ \hline
$x_{i}^{t}$       & congestion rate of the \textit{i}th road segment at time t                    \\ \hline
$\bar{v}_{i}^{t}, v_{i}^{t}$ & historical average speed and harmonic mean speed of the \textit{i}th segment at time t                \\ \hline
$\hat{v}_{i}$     & free-flow speed  of the \textit{i}th segment                                  \\ \hline
$\tau$            & the length of a sliding window                                                \\ \hline
$t_0$             & the start time step of the prediction window in a sliding window              \\ \hline
$B$               & a batch of sliding windows                                                    \\ \hline
$e$               & encoding vector produced by Encoder for a specific context window             \\ \hline
$h^t$             & LSTM hidden states at time t                                                  \\ \hline
$K$               & the number of Coupling layer and Batch normalizing layer in the RealNVP model \\ \hline
$z$               & the base distribution of the flow model                                       \\ \hline
$y$               & an arbitrary latent representation learned as part of the transformation in flow model \\ \hline 
$p(x^t)$          & probability density of $x^t$                                                  \\ \hline
$s_{net}, t_{net}$      & \textit{scale} and $translation$ network in each Coupling layer of the RealNVP model                         \\ \hline
$history(x^j)$       & history data that are at the same period as $x^j$, where $j\in[t_0, \tau]$ is a detected anomalous time step \\ \hline
$\theta$          & trainable parameters in the model                                             \\ \hline
$\alpha, \beta$          & the fractions of anomalous time slices and road segments in the synthetic dataset                \\ \hline
\end{tabularx}
\caption{List of Symbols}
\label{tab:symbols}
\end{table}

Let $S$ denote the set of all road segments under consideration. Consider an arbitrary segment $S_i \in S$ on which (near) real-time speed is monitored continuously; we assume that the estimated harmonic mean speed on segment $S_i$ is computed and stored at discrete times $t\in \{1, 2, ..., T\}$. We denote this observation at time $t$ by $v_{i}^t$. The free-flow speed $\hat{v}_{i}$, an intrinsic property of segment $S_i$, is calculated based upon the 85th-percentile of the observed speeds on the segment $S_i$ for all time periods~\cite{dixon1999posted,park2013quantifying,pankaj2019comprehensive}. The historical average speed, denoted by $\bar{v}_{i}^{t}$, signifies the regular traffic condition on $S_i$, which is calculated by taking the harmonic average of speeds on $S_i$ for each hour of day and for each day of the week. Then, the congestion rate is defined as
\begin{equation}
\label{eq:congestion}
    x_{i}^{t} = \frac{\bar{v}_{i}^{t}-v_{i}^{t}}{\hat{v}_{i}}
\end{equation}

The congestion rate of $N$ roadway segments can be modeled as an $N$-dimensional time series of length $T$, denoted by $X$, i.e., $X=\{x^1,x^2,\dots x^T\}$, where $x^t\in \mathbb{R}^{N}$ is an $N$ dimensional vector representing a measurement at time $t$. The congestion observation from the $i$th segment at time $t$ is $x^t_i$. At an arbitrary time $t$, $x^t$ therefore denotes a snapshot of congestion at all roadway segments. Each time step can have additional features associated with it, e.g., day of the week and hour of the day. We denote such features for the $t$-th time step be $\lambda^t$. 

The primary goal of our framework is to detect points in time at which anomalous congestion may occur. With the obtained detection results, the secondary target is to recognize the roadway segments most likely to have caused the abnormal observation at each timestamp. Figures~\ref{fig:clustering} and~\ref{fig:model} together demonstrate a four step method we propose for fulfilling these targets: (1) time series clustering based on similarity measures; (2) unsupervised anomaly detection based on conditional RealNVP; (3) anomaly diagnosis at the road segment granularity based on non-parametric kernel density estimation; (4) auxiliary supervised anomaly detection based on multi-layer perceptron.

\subsection{Time Series Clustering}\label{subsec:clustering}
In practice, $X$ might be composed of thousands of dimensions with heterogeneous temporal patterns, semantic meanings, or underlying dependencies. It is computationally difficult to learn patterns or explicit probability distributions for extremely high-dimensional data. One way to alleviate this challenge is by identifying dimensions that are related in the feature space. To tackle this, we perform data-driven clustering to partition the given time series into separate groups based on similarity (where similarity is based on an appropriate distance in the feature space, e.g., the $\ell1$ norm). This step facilitates anomaly detection and diagnosis in two aspects. First, it ensures that learning the probability distribution over the input time series is tractable. Second, similarity in the feature space naturally associates semantic meaning to the clusters; e.g., we observe that different clusters correspond (roughly) to different types of roads such as highways and on-ramps. Learning an explicit distribution for a particular cluster therefore enables us to learn a distribution of traffic in a particular type of roadway.
\begin{figure}[htb]
\centering
\includegraphics[width=\columnwidth]{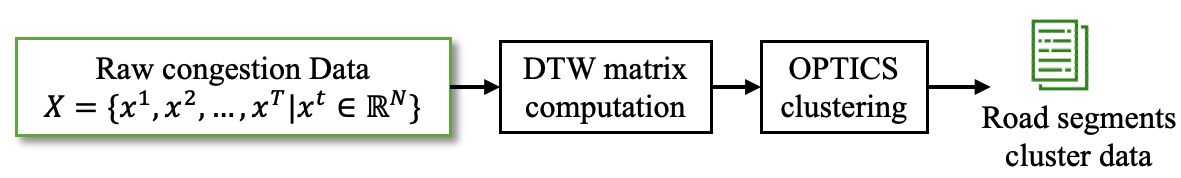}
\caption{Grouping the road segments into clusters.}
\label{fig:clustering}
\end{figure}

Traditional clustering methods cannot be applied directly to time-series settings due to the temporal nature of the data. A general idea of time series clustering, as shown in Figure~\ref{fig:clustering}, is to first convert temporal data to a \textit{flat} representation by computing a similarity or distance matrix; then,  a standard clustering algorithm (e.g., KMeans~\cite{lloyd1982least}, DBSCAN~\cite{schubert2017dbscan}) can be used to partition the flat representation. We leverage the commonly-used ``Dynamic Time Warping'' (DTW)~\cite{berndt1994using} distance to measure pair-wise similarities between time series. Particularly, consider arbitrary road segments $S_i$ and $S_j$. The DTW distance between the segments (in the feature space defined by congestion) can be calculated as the squared root of the sum of squared distances between every element in $x^t_i \; \forall t \in \{1,\dots,T\}$ and its nearest point in $x^t_j \; \forall t \in \{1,\dots,T\}$.  Intuitively, the distance reflects how similar was \textit{each} congestion value observed in segment $S_i$ to \textit{any} congestion value observed in segment $S_j$, and then aggregates the similarities. Given the similarity measures, we use the OPTICS algorithm~\cite{ankerst1999optics} for clustering. The OPTICS algorithm is density-based, which does not require the number of clusters as a prerequisite. Given the clusters, we perform anomaly detection in each cluster independently.

\subsection{Timestamp-level Anomaly Detection}

\begin{figure}[t]
\centering
\includegraphics[width=\columnwidth]{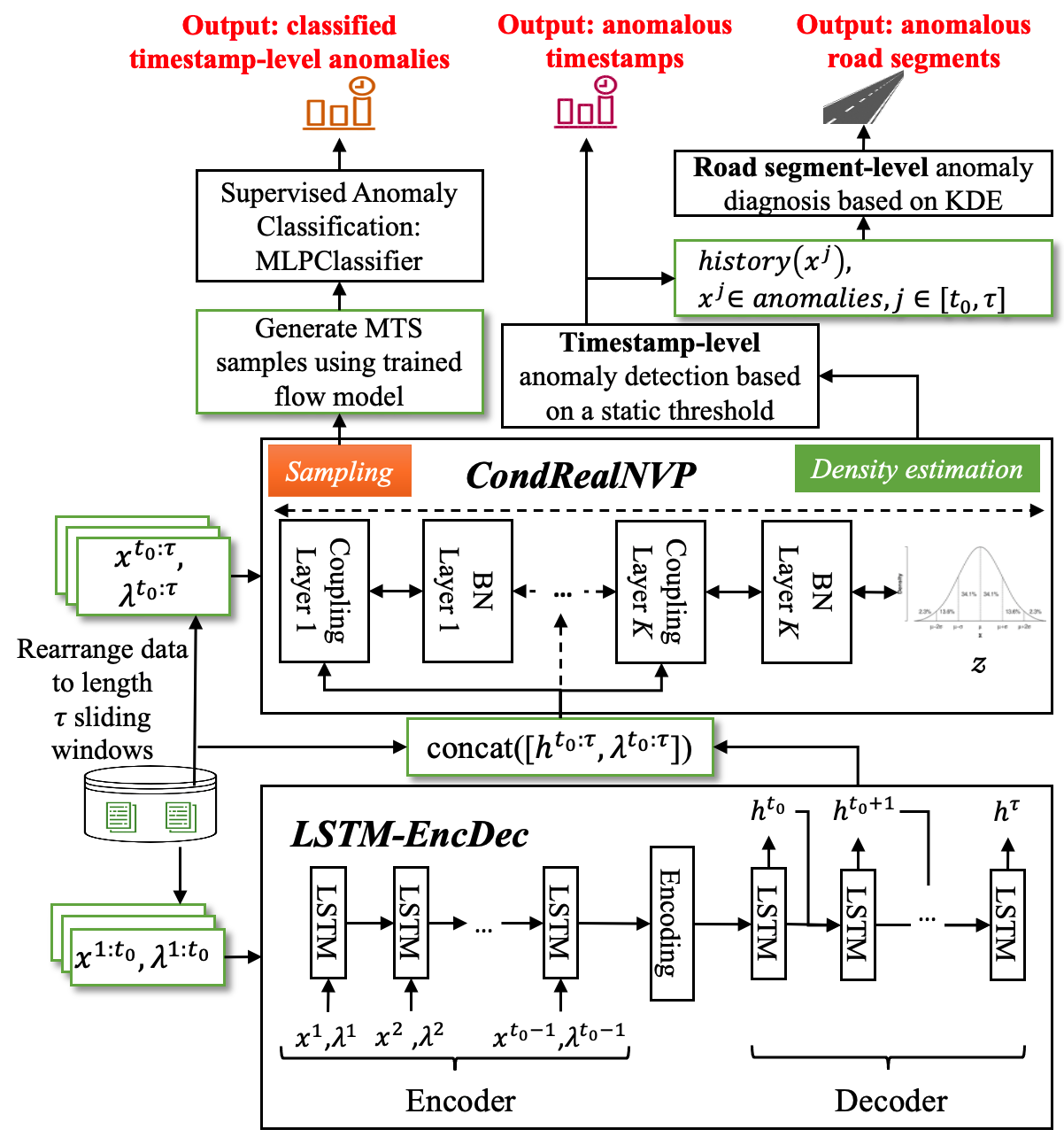}
\caption{Anomaly detection, roadway segment-level anomaly diagnosis, and supervised anomaly classification.}
\label{fig:model}
\end{figure}

We can perform density estimation on the raw data using RealNVP directly; recall that the normalizing flow approach allows us to perform tractable density estimation. However, our initial experiments proved otherwise as the flow model failed to capture contextual anomalies. This observation is not surprising; the transformations in RealNVP operate along the feature dimensions but discard the temporal correlation in the data. In contrast, recurrent neural networks with gated memory such as LSTM (long short-term memory networks)~\cite{hochreiter1997long} 
have been proven to be powerful tools for modeling sequential data. This inspires us to explore the possibility of capturing ``point'' and ``contextual'' anomalies simultaneously by aggregating an LSTM-based Encoder-Decoder and a normalizing flow model. As LSTM requires three-dimensional inputs (the batch size, the number of time steps, and the number of features), we use overlapping sliding windows $x^{\{1:\tau\}}$ of length $\tau$ as inputs, where each window is further divided into a context window $x^{\{1:t_{0}-1\}}$ and a prediction window $x^{\{t_{0}:\tau\}}$ (see figure~\ref{fig:model}). We explain the functioning of the LSTM below.


\subsubsection{LSTM Encoder-Decoder} We use an LSTM-Encoder-Decoder structure that defines two separate components: an encoder and a decoder, each of which is composed of a stack of LSTM layers. During inference, the encoder first converts data $x^{\{1:t_0-1\}}$ into a single fixed-length representation vector, $f_{\theta_{enc}}: x^{\{1:t_0-1\}}\rightarrow e$, given by the last hidden states of LSTM, that contains all the information needed for the input of a subsequent decoder. The encoder vector $e$ is then repeated $\tau-t_0+1$ times and used to initialize the internal states of decoder LSTM cells. The decoder then generates the ultimate hidden states of the target sequence $x^{t_0:\tau}$ in an autoregressive manner, i.e., $f_{\theta_{dec}}: e\rightarrow h^{t_0:\tau}$.

With respect to anomaly detection, the autoregressive scheme enables RNN to propagate and leverage historical information. Also, the encoder-decoder structure prevents out-of-distribution data from being constructed from compressed historical information, which in turn ensures that the density is learned without too many outliers. Deterministic encoder-decoder models use mean squared error as anomaly score to measure the deviation between observations and predictions. However, this score may result in sub-optimal anomaly detection decisions due to two reasons. First, noise in data and randomness from the model's parameters may interfere with the training procedures. Second, the decisions are concluded from a fixed-length of historical data, therefore lacking a global perspective. We bypass this issue by integrating the encoder-decoder structure with the normalizing flow (described below) and training the overall architecture by maximizing the log likelihood function (which is inherently probabilistic). 

Consider an observation $x^t$ where $t \in \{t_0:\tau\}$. Let the output of the last layer of the decoder for $x^t$ be denoted by $h^t$. Intuitively, given $h^t$, which implicitly contains information summarized from previous time steps, our ultimate goal is to estimate the likelihood of $x^t$ in the entire input space $\mathcal{X}$, i.e., $p(x^t|h^t)$. A low value implies the observation is either rare in the input space or deviates from contextual behavior. Our implementation of the LSTM Encoder-Decoder model is shown in figure~\ref{fig:model}. We include temporal features such as week-of-year, day-of-week, and hour-of-day with the encoder's input to facilitate learning the seasonality and trend patterns of time series. Then, the likelihood of the observations from $t_0$ to $\tau$ can be represented as:

\small
\begin{align}\label{eqn:density}
p(x^{t_0:\tau} \mid x^{1:t_0-1}, \lambda^{1:t_0-1}, \theta_{enc},\theta_{dec}) &= \prod_{t=t_0}^{\tau}p(x^t|h^t,\lambda^t,\theta_{dec}) 
\end{align}
\normalsize
where $h^t$ denotes the LSTM hidden at $x^t$ that is autoregressively derived from the previous step.
Next, we explain how to compute the density function mentioned in  equation~\ref{eqn:density} using a conditional RealNVP flow model.

\subsubsection{Conditional RealNVP}
Note that while the LSTM requires a three-dimensional input, such an input cannot directly fit into the flow model. As a result, we begin by flattening the time dimension. Recall that our goal is to learn a set of bijective functions that enable transformation between a simple distribution and the real-world data distribution, as mentioned in section~\ref{subsec:realnvp}.  We use a multivariate Gaussian distribution with a diagonal covariance matrix as the base distribution, which is a common choice for normalizing flow~\cite{kobyzev2020normalizing}. Let $y^t$ denote an arbitrary latent representation learned as part of the transformation. 
RealNVP partitions a given $x$ into two disjoint groups, one of which is unchanged and mapped to the $1:d$ dimensions, while the other part of $x$ undergoes a transformation and is mapped to $d+1:D$ dimensions (see equation~\ref{eqn:transformation} in section~\ref{sec:background}). 
To model the conditional distribution shown in equation~\ref{eqn:density}, we concatenate $h^t$ and $\lambda^t$ with the unchanged part of $y^t$, forming the inputs of \textit{st-networks} in each coupling layer (see figure~\ref{fig:model}). During the transformations, we use binary mask vectors to extract the changed and unchanged dimensions in $y^t$, where the unchanged dimensions are multiplied by ones and the other dimensions are multiplied by zeros. Note that the outputs of \textit{st-networks} preserve the dimensions of $d+1:D$ using the inverse mask vector so that we can compute corresponding values smoothly. 


We stack $K$ coupling layers to ensure the flow models can perform adequate changeovers when modeling complicated distributions, corresponding to the second term in equation~\ref{eqn:nvp_density_estimation}. We also place a bijective batch normalization (BN) layer after every coupling layer. Our design is motivated by prior work by Dinh et.al~\cite{dinh2016density}, who use BN layers to stabilize the training process. As the BN layer is essentially a linear function, it is invertible and the computation of the Jacobian is efficient. 

\subsubsection{Training and Inference}
The flow and encoder-decoder models are trained together via minimizing the below loss function with the Adam~\cite{kingma2014adam} optimizer. Given a batch of sliding windows $B$, according to the optimization objective~\ref{eqn:nvp_density_estimation} and equation~\ref{eqn:density}, the loss function is parameterized as:

\small
\begin{equation}
    \mathcal{L} = -\frac{1}{|B|\cdot (\tau-t_0+1)} \sum_{x^{t_0:\tau} \in B}\sum_{t=t_0}^\tau \log{p_{X}}(x^t|h^t, \lambda^t; \theta)
\end{equation}
\normalsize
where $\theta$ denotes all trainable parameters in the workflow.

During inference, the procedure of anomaly detection is straightforward and computationally efficient after training. Given a sample $x^t$, we use the trained network to perform density estimation, and flag the point as an anomaly based on a exogenous threshold $\epsilon$.


\subsection{Segment-level Anomaly Diagnosis}
We now have a general architecture that can detect anomalies from real-time congestion data. However, note that an anomalous data point $x^t$ (say) consists of $N$ dimensions, where each dimension corresponds to a road segment. This detection does not fully solve our problem; recall that our goal is to enable TMC operators focus their attention (e.g., secondary inspection of cameras and resource allocation) to a small subset of segments. However, $N$ can still be large in practice. Now, given an anomalous time vector $x^t$, we describe how to diagnose an anomaly down to the granularity of an individual segment.
Consider $x^t=<x_1, x_2, ..., x_n>$ is a detected anomalous vector, we investigate the data distribution at time $t$ by gathering historical data at the same period of $[t-\frac{\sigma}{2}, t+\frac{\sigma}{2}]$, where $\sigma$ denotes a configurable window size. Then, we train a density estimation model with a Gaussian kernel for each time series and determine the density threshold using a split validation dataset. 

\subsection{Supervised Anomaly Classification}
Given the normalizing flow model (that can perform exact density estimation and efficient sampling) and the LSTM-EncDec model (that can capture temporal correlations), we can generate labeled synthetic data to train supervised classifiers for anomaly detection. The procedure of generating MTS sequences are as follows: we first provide a warm-up sequence (an initial context window) as the input of the encoder to produce the decoder's initial hidden states. Anomalies and normal samples are then sampled from a standard normal distribution and then transformed to the output space (with decoder hidden states as conditional inputs). Generated samples are reused as inputs of the next iteration until the desired time series length is reached. The samples can then be used to train a classifier. We use a multi-layer perceptron classifier in our analysis.

\subsection{Thresholding}
CondRealNVP and baseline methods label anomalies by comparing the computed anomaly score (probability density, reconstruction, or prediction errors) with a static or dynamic threshold. In this project, we examine a static thresholding method and a dynamic one. The static method iteratively searches anomaly scores of the first $N\%$ data points of the testing dataset for a threshold that yields the maximum F1-Score. The computed threshold is then applied to anomaly scores of the remaining data. We utilize the SPOT~\cite{siffer2017anomaly} algorithm for dynamic thresholding, which essentially combines the Peak Over Threshold (POT) algorithm with a streaming process. Specifically, POT is a statistical method that uses “extreme value theory” to fit the data distribution with a Generalized Pareto Distribution. SPOT algorithm begins with computing an initial threshold by performing a POT estimate on the first $N$ data points. SPOT then (1) updates its model using incoming values that are smaller than the threshold; (2) and fits values greater than the threshold to a GPD to adjust the threshold.

\section{Evaluation}
In this section, we first introduce a real-world traffic dataset and a public server machine dataset used to train and evaluate our approach. For the traffic dataset, we (1) report the anomaly detection and diagnosis performance on synthetic datasets in terms of Recall and F1-score; (2) present the correlations between inferenced anomaly scores with real traffic anomalies; (3) demonstrate the anomaly classification performance of the CondRealNVP-driven supervised classifier in terms of AUC score. For the server machine dataset, we present the timestamp-levle anomaly detection performance when integrating our approach with a dynamic thresholding method. A thorough comparison with several state-of-the-art anomaly detectors was also conducted to illustrate the effectiveness of our approach. 

\subsection{Datasets}
\subsubsection{Traffic Data} We use an INRIX\footnote{\url{https://inrix.com/}} traffic mobility data collected for one year (2019) from the city of Nashville, Tennessee. This dataset contains estimated ``real-time'' harmonic mean flow speeds, free-flow (reference) speeds, and historical average speeds of 364 interstate road segments with a five-minute frequency. The congestion rate measurements are derived based on the equation~\ref{eq:congestion}. We impute missing values at a specific road segment by interpolating observations from nearby segments. If nearby segments also contain missing values, we impute by using historical averages.

\begin{table}[htb]
\centering
\resizebox{\columnwidth}{!}{%
\begin{tabular}{cc}
\hline
Property            & Values                                  \\ \hline
\# roadway segments & 364                                     \\
\# records/segments & 104832                                  \\
collection period              & 2019-01-01 00:00 --2019-12-30 23:55 \\
frequency           & 5 minutes                               \\ \hline
\end{tabular}
}
\caption{Details of the traffic dataset collected from Nashville TN}
\label{tab:congestion-data-table}
\end{table}

\textit{Synthetic Testing Data} Note that the real-world traffic data is unlabeled. To evaluate the efficacy of the proposed approach, we generate synthetic data. We use the actual congestion data from Oct-2019 to Dec-2019 to generate this dataset. First, we model the ground-truth MTS as a multivariate Gaussian distribution, whose parameters are learned from empirical observations. We sample from the multivariate Gaussian. Then, we randomly inject ``point'' and ``contextual'' anomalies at a fraction ($\alpha$) of half-hour length time slices and in a fraction ($\beta$) of road segments respectively, where $\alpha$ and $\beta$ are hyper-parameters that control the temporal and spatial distribution of anomalies. The motivation of injecting anomalies in a temporal manner is that traffic congestions are not instantaneous events in practice. Point anomalies are created by perturbing the values obtained from the first step by a factor drawn from a uniform distribution $\mathcal{U}(-g, +g)$, where $g$ denotes the magnitude of congestion rate of the day. Contextual anomalies are introduced by flipping the time slices that have minimum and maximum hourly average values. 

\subsubsection{Server Machine Data}
SMD (Server Machine Dataset)~\footnote{~\url{https://github.com/NetManAIOps/OmniAnomaly}} is a public 5-week-long dataset which was collected by OmniAnomaly~\cite{su2019robust} authors from a large Internet company with a one-minute frequency. The dataset is divided into three groups based on the service hosted on machines. There are 28 machines in total and 38 KPIs are collected on each machine. The SMD dataset is divided into two subsets of equal size: the first half is the training set and the remaining is the testing set. Anomalies and their anomalous dimensions in SMD testing set have been labeled by domain experts based on incident reports. The details of the SMD dataset is summarized in Table~\ref{tab:smd-data-table}.

\begin{table}[htb]
\centering
\resizebox{\columnwidth}{!}{%
\begin{tabular}{cc}
\hline
Property            & Values                                  \\ \hline
\# data files & 28                                     \\
\# dimensions & 38                                  \\
frequency     &    1 minute \\
training set size             & 708405 \\
testing set size           & 708420                          \\
anomaly ratio(\%)   & 4.16 \\
KPIs             & CPU load, network usage, memory usage, etc. \\
\hline
\end{tabular}
}
\caption{Details of the SMD dataset}
\label{tab:smd-data-table}
\end{table}

\subsection{Baselines}
In terms of MTS anomaly detection, we compare our approach against prediction-based (e.g., AR, DeepLog-LSTM) and reconstruction-based (e.g., AE, VAE, EncDec-AD) anomaly detectors. These models are briefly described as the following:

\begin{enumerate}
    \item AR~\cite{lai2020tods} models use linear regression to calculate a sample’s deviance from the predicted value, which is then used as its outlier scores. This model is for multivariate time series. This model handles multivariate time series by various combination approaches. Specifically, this algorithm trains independent linear regression models for each dimension then computes the anomaly score for every sample based on the mean, maximization, or median of weighted deviance of each dimension;
    \item AE~\cite{hawkins2002outlier} is an autoencoder model based on MLP that compresses data using an encoder and decode it to retain original structure using a decoder. AE could be used to detect outlying objects in the data by calculating the reconstruction errors; 
    \item VAE is a variational autoencoder with MLP encoder and decoder;
    \item EncDec-AD~\cite{malhotra2016lstm} replaces MLP layers in AE with LSTM layers;
    \item DeepLog-LSTM~\cite{du2017deeplog} is a deep neural network model that utilizes LSTM to model a system log as a natural language sequence. This allows DeepLog to automatically learn log patterns from normal execution, and detect anomalies when log patterns deviate from the model trained from log data under normal execution. TODS library~\cite{lai2020tods} only implements the Parameter Value anomaly detection model in DeepLog~\cite{du2017deeplog} for time series data, which essentially realizes a stacked LSTM model.
\end{enumerate}
As the baselines do not support anomaly diagnosis at the feature level, we adopt the same KDE-based method as in the proposed method when performing evaluation on the Inrix dataset.

\subsection{Model Configurations}
For experiments on the traffic dataset, we configure the encoder and decoder of LSTM-ED model to have 2 LSTM layers. The flow model consists of 10 interleaving bijection and Batch Normalization layers. For both dataset, the $s_{net}$ of the \textit{st-network} is activated with the Tanh function, and the $t_{net}$ uses the ReLU function~\cite{goodfellow2016deep}. The model is trained for a maximum of 300 epochs with a batch size of 64. Out of the training set, 30\% is kept out as validation set for early stopping. We performed hyper-parameter tuning for experiments on both datasets. More detailed hyper-parameter settings of each model are reported in Appendix~\ref{app:model_config}.

All experiments are run on a single Nvidia TITAN X GPU (12GB) and the code implementations are based on the Tensorflow Keras library version 2.4.0 and Tensorflow Probability 0.11.0. Our implementation can be found at \url{https://github.com/scope-lab-vu/CondRealNVP}.

\subsection{Experiment Setup}
\subsubsection{Traffic Data}
The high dimensionality of the MTS make computing the DTW distance matrix (see section~\ref{subsec:clustering}) time-consuming. Therefore, we sample 100 segments uniformly at random use one week of data from the segments to generate the cluster prototype. Then, we calculate the DTW distances between the remaining segments and centroids of initialized clusters and merge them into the nearest cluster. 

Our model is trained with the congestion data from Jan-2019 to Sep-2019 and evaluated on synthetic data of the remaining months. We empirically configure the sliding window size to 72 (6 hours) and the moving step length to 12 (1 hour). Point and contextual anomalies are detected together for all experiments. We evaluate the anomaly detection performance from two perspectives: effectiveness (based on temporal parameter ($\alpha$) and sensitivity (based on spatial parameter ($\beta$). Effectiveness measures whether anomalies can be found in the case of high imbalance between anomalous and normal data. Sensitivity, on the other hand, evaluates situations in which only a portion of road segments are under anomalous congestion at a specific time, which challenges our approach to capture anomalies with high sensitivity. For each $\alpha$ and $\beta$ pair, we generate the synthetic test set five times and calculate the average model performance. 

For each individual cluster, we generate five one-month long datasets. Normal and anomolous data are sampled from the overall flow architecture. Then for each cluster, we train an MLP classifier with the binary cross-entropy loss and Adam optimizer. Trained MLP classifiers are evaluated on synthetic datasets using the area under the curve (AUC) score metric.

\subsubsection{Server Machine Data}
Given the SMD dataset has been divided into three groups, it is unnecessary to perform time-series clustering. We selected one machine from each group (machine-1-1, machine-2-1, machine-3-7) and trained three individual models. Timestamp-level anomaly detection is then evaluated. Unlike the static thresholding approach employed on the traffic dataset, we leverage the SPOT algorithm to decide the threshold of labeling anomalies. Each model is trained and evaluated five times. We compare the performance between CondRealNVP and baselines in terms of F1-Score and Recall. 

\subsection{Results and Discussion for Traffic Data}
\textbf{Clustering} The MTS clustering step groups the 364 road segments into eight clusters that include 55, 67, 68, 79, 10, 12, 62, and 11 road segments, respectively. We name the clusters using the letters \textit{A-H}. Empirical results show that roadways in the same cluster usually have similar functions or properties. For instance, Cluster A mainly covers Exit road segments. Cluster B involves the highways (e.g., I-65, I-40) that connect Nashville towards neighboring cities. Cluster C consists of road segments around on-ramps. Experimental results for individual clusters can be found in our Github repository. 

\textbf{Anomaly Detection} First, we compare CondRealNVP with baseline methods for anomaly detection from aspects of effectiveness and sensitivity. The former is achieved by configuring the fraction of abnormal time slices $\alpha$ to 5\%, 3\%, and 1\%, and the latter is by setting the fraction of anomalous road segment $\beta$ to 100\%, 50\%, and 25\%. We conducted controlled experiments and fixed $\beta$ to 50\% when testing the effectiveness and configure $\alpha$ as 5\% for the sensitivity test. These settings ensure the sparsity of irregular traffic congestion in temporal and spatial. The effectiveness test results shown in Table~\ref{tab:effectiveness} reflect that CondRealNVP consistently outperforms other methods, with average improvements of 0.203--0.335 and 0.154--0.212 in terms of average Recall and F1-Score. There is a clear trend that the Recall and F1-Score degrade as the decreasing $\alpha$; however, our approach is relatively more robust and guarantee acceptable performance even in the case of $\alpha=1\%$. We also observe similar results in the sensitivity analysis. Except the situation where all road segments in a cluster suffer heavy congestion in a specific time slice ($\beta=100\%$), we observe that the \textit{CondRealNVP} model comprehensively outperforms the other approaches.

\begin{table}[htb]
\vspace{-5pt}
\centering
\resizebox{0.9\columnwidth}{!}{%
\begin{tabular}{l|ccc|ccc}
\hline
\multicolumn{1}{c|}{metrics}                 & \multicolumn{3}{c|}{Recall}     & \multicolumn{3}{c}{F1-Score}  \\ \hline
\multicolumn{1}{c|}{anomaly rate $\alpha$} & 5\%    & 3\%    & 1\%    & 5\%    & 3\%    & 1\%    \\ \hline
AR                                        & 0.376 & 0.343 & 0.265    &   0.446 & 0.401 & 0.292  \\
AE                                        & 0.576 & 0.504 & 0.359    &    0.475 & 0.421 & 0.295  \\
VAE                                       & 0.574 & 0.518 & 0.362    &   0.490 & 0.435 & 0.307 \\
EncDec-AD                                 & 0.529 & 0.472 & 0.358    &   0.466 & 0.401 & 0.272  \\
DeepLog-LSTM                              & 0.411 & 0.345 & 0.278    &   0.439 & 0.374 & 0.246  \\
CondRealNVP                               & \textbf{0.752} & \textbf{0.710} & \textbf{0.604}    &   \textbf{0.632} & \textbf{0.583} & \textbf{0.480}   \\ \hline
\end{tabular}%
}
\caption{Avg. Recall and F1-score of the effectiveness test across 8 clusters under different anomaly rates ($\beta=50\%$). Best results are presented in bold.}
\label{tab:effectiveness}
\end{table}

\begin{table}[htb]
\vspace{-5pt}
\centering
\resizebox{0.9\columnwidth}{!}{%
\begin{tabular}{l|ccc|ccc}
\hline
\multicolumn{1}{c|}{metrics}                 & \multicolumn{3}{c|}{Recall}     & \multicolumn{3}{c}{F1-Score}  \\ \hline
\multicolumn{1}{c|}{road segments $\beta$} & 100\%    & 50\%    & 25\%    & 100\%    & 50\%    & 25\%    \\ \hline
AR                                        & 0.482 & 0.376 & 0.275    &   \textbf{0.580} & 0.446 & 0.241      \\
AE                                        & 0.446 & 0.576 & 0.504    &   0.516 & 0.475 & 0.369   \\
VAE                                       & 0.459 & 0.574 & 0.486    &   0.541 & 0.490 & 0.370     \\
EncDec-AD                                 & 0.468 & 0.529 & 0.462    &   0.519 & 0.466 & 0.342     \\
DeepLog-LSTM                              & 0.429 & 0.411 & 0.359    &   0.530 & 0.439 & 0.288     \\
CondRealNVP                               & \textbf{0.530} & \textbf{0.752} & \textbf{0.648}    &   0.575 & \textbf{0.632} & \textbf{0.484}   \\ \hline
\end{tabular}%
}
\caption{Avg. Recall and F1-Score of sensitivity test under different ratios of anomalous road segments ($\alpha=5\%$).}
\label{tab:sensitivity}
\end{table}

\textbf{Segment-Level Detection} Given cluster level anomaly detection, we now evaluate the accuracy of performing segment level detection. We present the experimental results in Table~\ref{tab:diagnosis}. It can be seen that the overall trend coincides with what we observed in the sensitivity test. Specifically, CondRealNVP obtains 0.102--0.184 and 0.031--0.078 improvement regarding Recall and F1-Score compared with baseline methods. Finally, recall that anomaly detection for traffic centers is intended in near real-time. On average, for each cluster, inference (including the time taken to train KDE models) takes around 34 milliseconds, which is an acceptable latency in practice.

\begin{table}[htb]
\vspace{-5pt}
\centering
\resizebox{0.9\columnwidth}{!}{%
\begin{tabular}{l|ccc|ccc}
\hline
\multicolumn{1}{c|}{metrics}                 & \multicolumn{3}{c|}{Recall}     & \multicolumn{3}{c}{F1-Score}  \\ \hline
\multicolumn{1}{c|}{road segments $\beta$} & 100\%    & 50\%    & 25\%    & 100\%    & 50\%    & 25\%    \\ \hline
AR                                        & 0.459 & 0.238 & 0.126    &   0.570 & 0.341 & 0.178      \\
AE                                        & 0.401 & 0.322 & 0.210    &   0.491 & 0.345 & 0.221   \\
VAE                                       & 0.422 & 0.333 & 0.211    &   0.517 & 0.352 & 0.218    \\
EncDec-AD                                 & 0.427 & 0.300 & 0.192    &   0.507 & 0.341 & 0.211     \\
DeepLog-LSTM                              & 0.395 & 0.242 & 0.156    &   0.514 & 0.333 & 0.199    \\
CondRealNVP                               & \textbf{0.520} & \textbf{0.508} & \textbf{0.282}    &   \textbf{0.572} & \textbf{0.420} & \textbf{0.245}   \\ \hline
\end{tabular}%
}
\caption{Avg. Recall and F1-Score for segment-level anomaly diagnosis with $\alpha=5\%$.}
\label{tab:diagnosis}
\end{table}

\begin{figure*}[]
\centering
\subfigure{
    \label{fig:congestion_rate}
    \includegraphics[width=\columnwidth,height=3cm]{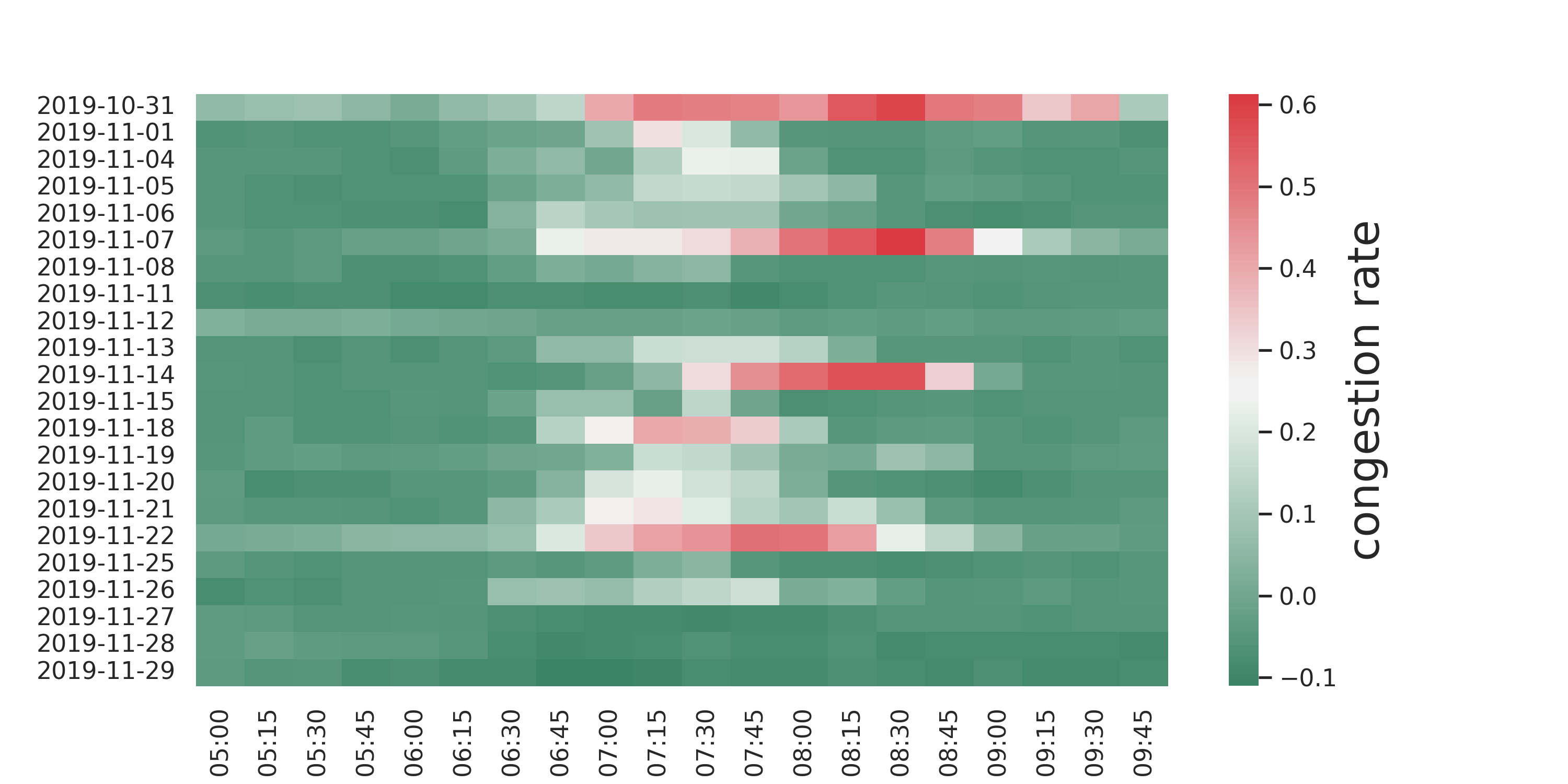}
}%
\subfigure{
    \label{fig:anomaly_score}
    \includegraphics[width=\columnwidth,height=3cm]{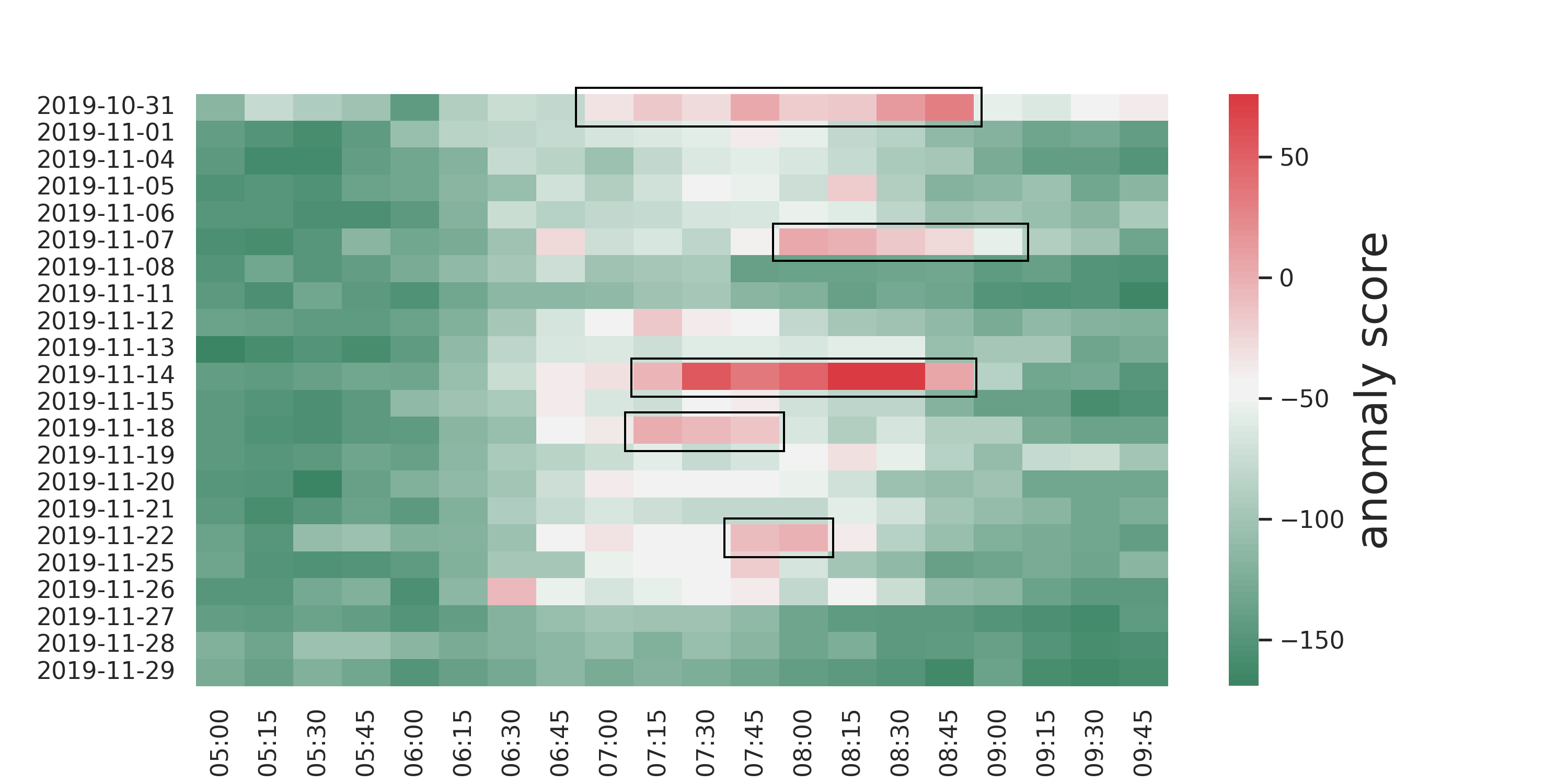}
}%
\caption{Visualization of real-world traffic anomalies and anomaly scores. The top figure shows the 85th percentile congestion rate (average in 15 minutes). The bottom figure shows the average anomaly scores per 15 minutes. Boxes highlight the most noticeable periods that are likely under abnormal congestion. Heatmaps show the results of Cluster G from 5 AM to 10 AM (rush hours) for 22 working days. Our method sensitively captures the time periods when recurring congestion occurred.}
\label{fig:congestion_visualization}
\end{figure*}

\textbf{Visualization of real-world traffic anomaly} We also show a case study on real-world data to evaluate our approach. We use real congestion data from cluster G for weekdays between Oct. 31st and Nov. 29th in Q4 2019, which is not seen during training. Figure~\ref{fig:congestion_visualization} shows (a) the 85th percentile congestion rate (average in 15 minutes) for all segments in the cluster and (b) the corresponding average anomaly score assigned by CondRealNVP. We select the period between 5 AM and 10 AM, which generally covers rush hours of the roadway (e.g., interstates 65) in Cluster G. The 85th percentile congestion intensity indicates the extent of congestion in the cluster, implying that only 15\% road segments are under heavier congestion states than presented. It can be seen that anomalous congestion occurred at 10/31, 11/07, 14, 18, and 22 with an apparent cascading pattern. As expected, our approach successfully discovers the peak hours and assigns notable anomaly scores from early stages.

\begin{figure}[]
\centering
\includegraphics[width=\columnwidth]{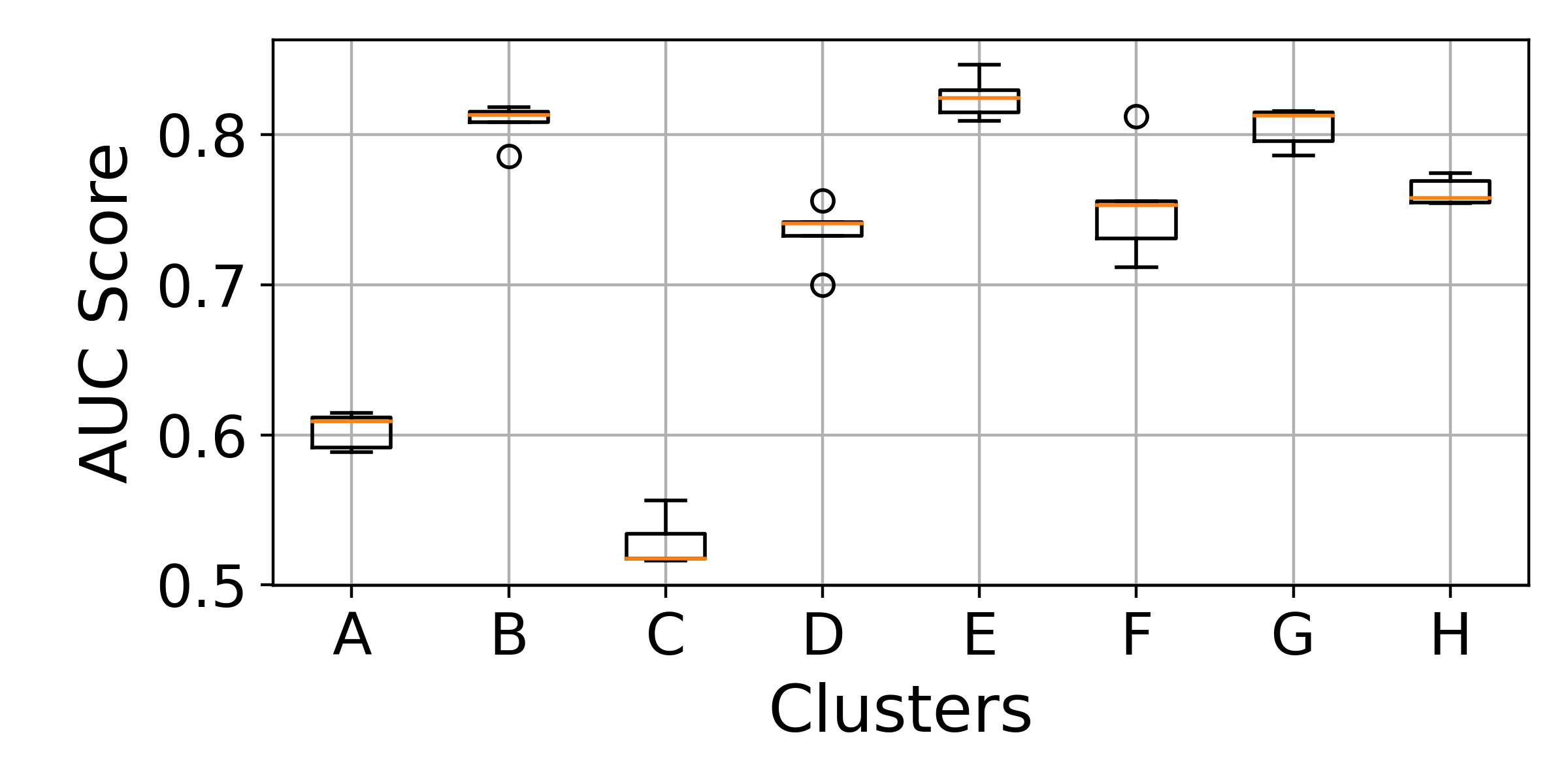}
\caption{Average AUC score for the MLPClassifiers on 5 synthetic testing datasets. Boxplot shows the performance variation of MLPClassifiers trained on 5 datasets drawn from CondRealNVP.}
\label{fig:classification}
\end{figure}

\textbf{Supervised Classification} Finally, we evaluate the efficacy of learning a classifiers in a supervised setting using samples drawn from CondRealNVP.  The classifiers are evaluated on five synthetic testing datasets, as we conducted in the previous section, with $\alpha=5\%$ and $\beta=50\%$. We report the average AUC score in Figure~\ref{fig:classification}. One can see that classifiers have acceptable discrimination capability (AUC score $\geq$ 0.7) in 6 of the 8 clusters. The relatively lower AUC scores in clusters A (off-ramp/Exit segments) and C (on-ramp segments) are probably due to the extremely low volume of abnormal congestion data in such clusters.

\subsection{Results and Discussion for Server Machine Data}
\begin{figure}[htb]
\centering
\includegraphics[width=\columnwidth]{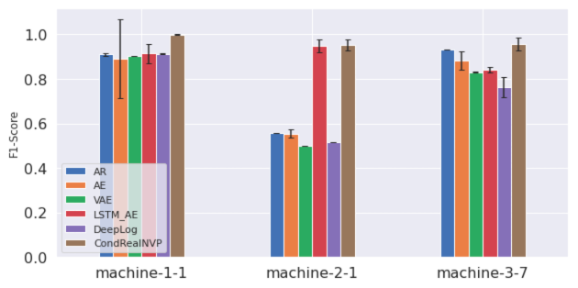}
\caption{Average F1-Score and standard deviation over five runs on the SMD dataset}
\label{fig:smd_f1}
\end{figure}

\begin{figure}[htb]
\centering
\includegraphics[width=\columnwidth]{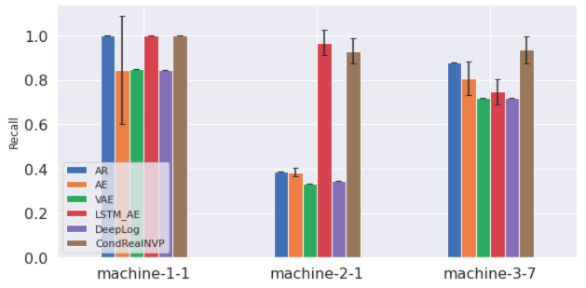}
\caption{Average Recall and standard deviation over five runs on the SMD dataset}
\label{fig:smd_recall}
\end{figure}

Figure~\ref{fig:smd_f1} indicates that CondRealNVP outperforms baselines by 0.068-0.239 on average over three selected machines in terms of F1-Score. The Recall of CondRealNVP, as shown in figure~\ref{fig:smd_recall}, is slightly lower than LSTM-AE on machine-2-1 but greater than baselines in other cases. Overall, CondRealNVP obtains an average of 0.050-0.323 higher recall over three selected machines.

\section{Related Work}
\label{sec:related}
Existing research on anomaly detection and diagnosis for surface transportation systems can be broadly classified into three classes according to the means of measuring deviations between normal samples and anomalies: \textit{reconstruction-based}, \textit{prediction-based}, and \textit{density-based} approaches. We review and investigate the advantages and weaknesses of each strategy and eventually propose CondRealNVP that combines the ideas of prediction-based and density-based approaches. 

\textbf{Reconstruction-based approaches} leverage the notion that normal samples can be better restructured from a compressed latent space than anomalies. The reconstruction error is used as anomaly score to indicate the deviation of anomalous samples from regular values. AutoEncoder and many of its variants have been the main foundation of this class of methods. Malhotra et al.~\cite{malhotra2016lstm} first introduced the LSTM encoder-decoder schema for time-series anomaly detection, known as EncDec-AD. Hu et.al~\cite{hu2020graph} combines AutoEncoder with graph convolutional networks to detect anomalies that lead to unexpected travel time in a set of directed weighted graphs. Madarash et.al~\cite{madarash2004enhancing} leverage LSTM predicted maneuver labels to reduce the number of false-positive alarms when using LSTM AutoEncoder for anomaly detection on driving modality data. A major limitation of these studies is that the detectors have to be built in a semi-supervised manner, training with unpolluted data only. This does not meet the unsupervised setting of our problem. Under the unsupervised setting, Chevrot et.al~\cite{chevrot2022cae} introduced \textit{Contextual AutoEncoder} to detect anomalies in En-Route trajectories. They improved the regular LSTM-based AutoEncoder using multiple decoders, where each focus on a specific flight phase (e.g. climbing, cruising, or descending). However, as shown in the study~\cite{pang2021deep}, a common issue with AutoEncoder-based methods is that the L2 optimization objective enforces models to learn a generic summarization of underlying regularities of the training dataset, even for outliers, leading to severe over-fitting. Variational AutoEncoder eliminates this problem by introducing an additional Kullback-Leibler divergence loss term. People therefore combined LSTM with VAE for multi-modal temporal anomaly detection in various domains~\cite{zhang2019velc, park2018multimodal, lin2020anomaly}.

\textbf{Prediction models} can naturally be used as anomaly detector, relying on the fact that normal samples are more predictable than anomalies. As the boosting of the complexity and dimension of realistic MTS data, memory-gated RNN models, such as LSTM and GRU, have been widely used on MTS forecasting and anomaly detection tasks due to their prominent capability in modeling long-term sequential dependencies. Malhotra et al.~\cite{malhotra2015long} first proposed LSTM-AD that uses stacked LSTM networks for MTS anomaly detection. Similar models are employed in DeepLog~\cite{du2017deeplog} for system log analysis. Channel-wise LSTM models are adopted by Hundman et al.~\cite{hundman2018detecting} for spacecraft anomaly detection. In terms of transportation systems, LSTM networks are applied on sensor data for traffic density estimation~\cite{nam2020deep} and speed forecasting~\cite{zhao2017lstm}. Chen et al.~\cite{chen2016long} leveraged stacked LSTM networks to predict traffic conditions with aids of four categories of traffic relevant online open data, including web map, social media, local events, and weather. Abduljabbar et al.~\cite{abduljabbar2021development} compared the performance of bidirectional LSTM (BiLSTM) and regular LSTM models for freeway traffic forecasting using simulation data. They highlight the superior performance of BiLSTM for multiple prediction horizons. Basak et al.~\cite{basak2019analyzing} analyzed the cascade effects of traffic congestion using a citywide ensemble of intersection level connected LSTM models and demonstrated a Timed Failure Propagation Graph-based diagnostics mechanism in ~\cite{basak2019data}. The major drawback of these attempts is that the forecasting accuracy is likely to be affected by anomalies when training models with polluted datasets~\cite{wu2020developing}, leading to unreliable anomaly detection results. In this case, the encoder-decoder structure was embedded in LSTM networks, known as sequence-to-sequence (Seq2Seq) prediction models. Loganathan et al.~\cite{loganathan2018sequence} leveraged the Seq2Seq encoder-decoder model to detect anomalous packets in TCP traffic. A more sophisticated Seq2Seq model with attention mechanism and bidirectional LSTM networks is introduced in~\cite{hao2019sequence} for short-term passenger flow prediction.

\textbf{Density-based approaches} intend to extract the underlying distribution of ground-truth data. The principle of detecting anomalies is that the density around a normal sample is similar to that around its neighbors. Chiang et al.~\cite{chiang2017btci} designed a two-step congestion cascades identification strategy, where they used a non-parametric Kernel Density Estimation function to compute anomaly score for road segments in the first step. Congested cascades are then formed by unifying both attribute coherence and spatio-temporal closeness of detected congested segments. Normalizing flows, as a class of recent probability density estimation technique, has been used for anomaly detection in transportation systems. Dias et al. ~\cite{dias2020anomaly} employed RealNVP and masked autoregressive flow for trajectory anomaly detection, in which a trajectory is defined as a sequence of GPS points generated by a moving object on a monitoring system. Their experimental results show that flow models outperform classical density-based methods including LOF~\cite{breunig2000lof} and a Gaussian mixture model~\cite{reynolds2009gaussian}. A remarkable benefit of density-based anomaly detection approaches is they do not need labeled training data~\cite{hodge2004survey}. However, since they only focus on the underlying data distribution, they cannot capture the sequential correlation in a time series. 

In addition to the natural limitations of the above studies, none of the existing work comprehensively addresses the scalability and spatial-temporal correlation challenges of anomaly detection and diagnosis for large traffic congestion datasets. In contrast, our presented work clusters road segments based on temporal similarity measures to decompose the problem scale. For an individual cluster, we employ the Encoder-Decoder-based LSTM model to extract sequential dependency and eliminate the interference of anomalies to the training process. A powerful conditional normalizing flow model then estimates the probability density of ground-truth congestion data. Real-time road segment-level anomaly diagnosis is also realized in our framework using fast kernel density estimation.
\section{Conclusion}
In this paper, we present an end-to-end framework to address the problem of multivariate time series anomaly detection and real-time anomaly diagnosis. In the framework, we identify similarity between uni-variate time series and use a conditional normalizing flow model for MTS that combines an LSTM Encoder-Decoder network with a RealNVP model for density-based anomaly detection. Then, we provide a KDE-based real-time anomaly diagnosis to locate anomalous features. Extensive experiments conducted on a real-world traffic dataset and a public server machine dataset manifest that our approach outperforms several state-of-art methods for both anomaly detection and anomaly diagnosis. 

In the future, for traffic anomaly detection, we plan to investigate the impact of more time-independent features, such as weather, holiday, events, etc. on anomaly detection performance. For machine KPI anomaly detection, we plan to improve the scalability of our approach to avoid training one model for every machine. In addition, we want to integrate attention mechanism with the LSTM networks to differentiate the importance of features and time steps when deriving anomaly score at a particular time. Finally, it is worth to explore other normalizing flow models, such as MAF, Glow, etc.

\appendices
\section{Model Configurations}
\label{app:model_config}
In this section, we document the model configurations of CondRealNVP when executing on the traffic and server machine datasets. For Clusters A-D and G of the traffic data, encoder LSTM layers consist of 128 and 64 hidden units (and the opposite for the decoder)~\cite{goodfellow2016deep}. The hidden size is changed to 64 and 32 for cluster E-F and H. The \textit{st-network} of every bijection layer is formed with 2 MLP layers (128 hidden dimensions for clusters A-D, G, and 32 for clusters E-F, H). We configure the context window size to 6 hours and 1 hour length for the prediction window. The moving step length is set to 1 hour. We summarize the hyper-parameter configurations of CondRealNVP when applying on the SMD dataset in Table~\ref{tab:smd_param}.

\begin{table}[htb]
\centering
\begin{tabular}{l|l|l|l}
           & machine-1-1 & machine-2-1 & machine-3-7 \\ \hline
window     & 360         & 1320        & 720         \\ 
steps      & 50          & 20          & 50          \\ 
\# bijection blocks & 15          & 5           & 20          \\ 
\# RNN units of LSTM-ED layers & 32          & 32          & 128         \\ 
\# RNN layers of LSTM-ED & 2           & 1           & 1           \\ 
\# units of \textit{st\_layers}   & 32          & 128         & 32          \\ 
\# \textit{st\_layer}  & 2           & 1           & 1           \\ 
learning rate         & 1e-4        & 1e-3        & 1e-4        \\ 
batch size & 128         & 128         & 64     \\    
\end{tabular}
\caption{Hyper-parameter settings of CondRealNVP when executing on SMD datset}
\label{tab:smd_param}
\end{table}

\balance
\bibliographystyle{IEEEtran}
\bibliography{references}

\end{document}